\documentclass[10pt,twocolumn,letterpaper]{article}

\usepackage{cvpr}
\usepackage{times}
\usepackage{graphicx}
\usepackage{epstopdf}
\usepackage{amsmath}
\usepackage{amssymb}
\usepackage{subcaption}
\usepackage{booktabs}
\usepackage{adjustbox}

\usepackage[pagebackref=true,breaklinks=true,letterpaper=true,colorlinks,bookmarks=false]{hyperref}

\cvprfinalcopy 

\def\cvprPaperID{14} 

\begin{document}

\title{Low-latency hand gesture recognition with a low resolution thermal imager}

\author{Maarten Vandersteegen$^1$, Wouter Reusen$^2$, Kristof Van Beeck$^1$ and Toon Goedem\'e$^1$\\
$^1$EAVISE, KU Leuven - Campus De Nayer - Belgium\\
{\tt\small \{firstname.lastname\}@kuleuven.be}\\
$^2$Melexis Technologies nv - Belgium\\
{\tt\small wre@melexis.com}
}


\maketitle

\begin{abstract}
    Using hand gestures to answer a call or to control the radio while driving a car, is
    nowadays an established feature in more expensive cars. High resolution time-of-flight
    cameras and powerful embedded processors usually form the heart of these gesture recognition systems.
    This however comes with a price tag. We therefore investigate the possibility to design an algorithm that predicts
    hand gestures using a cheap low-resolution thermal camera with only $32\times24$ pixels, which is light-weight
    enough to run on a low-cost processor.
    We recorded a new dataset of over 1300 video clips for training and evaluation and propose a light-weight 
    low-latency prediction algorithm. Our best model achieves 95.9\% classification accuracy and 83\% mAP detection
    accuracy while its processing pipeline has a latency of only one frame. 
\end{abstract}

\vspace{-1em}
\section{Introduction}

\begin{figure}
    \centering
    \includegraphics[width=.8\columnwidth]{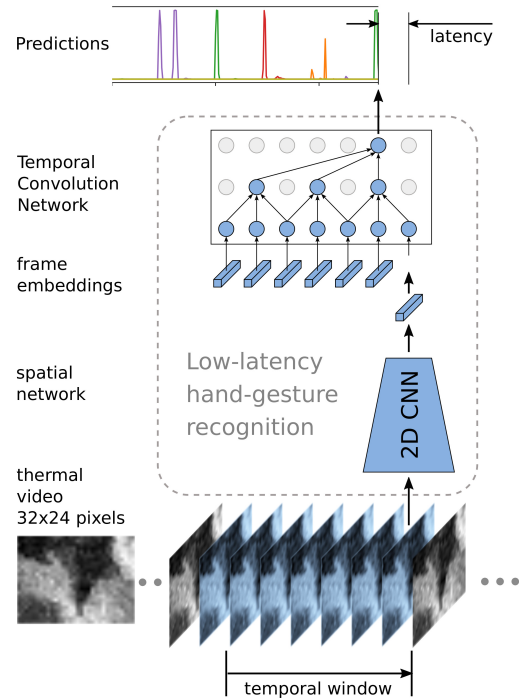}
    \caption{Proposed method for low-latency hand gesture detection. A 2D CNN summarizes frames into frame
             embeddings, which are processed by a 1D TCN with \textit{mixed causal} 
             convolutions. Only a single output close to the right edge of the temporal window produces the
             network's output prediction.}
    \label{fig:method_overview}
    \vspace{-1.5em}
\end{figure}

The human-machine interface controlled by hand-gestures is a well investigated research topic that targets many
different applications like controlling consumer electronics \cite{shan2010gesture}, operating the multimedia
system in a car \cite{Molchanov2016}, sign language interpretation \cite{jing2019recognizing} and even manipulating a
surgical robot \cite{vargas2014gesture}.
In the quest for improving the robustness and user-friendliness, different algorithms and sensors
have been proposed to perform the task at hand. Given the unpredictable light conditions in a car or an operating 
room, time-of-flight or structured IR-light based depth sensors are often preferred 
over regular color cameras for optical recognition applications. Also, given the nature of a depth image, the hand
performing the gesture close to the camera is easily separated from background clutter using a simple threshold.
These advantages however come with a price tag. Depth cameras are expensive, which is especially problematic for
low-cost applications.
Nowadays high-end cars have hand-gesture controllable multimedia systems which usually also rely on depth sensing
technology to capture the gesture information.
In this work, we will investigate the possibility to make a low-cost variant of an in-car human-machine interface
for controlling such a multimedia system using the low-cost MLX90640 thermal sensor from Melexis.
The MLX90640 is a thermopile imager array with 32$\times$24 pixels and configurable frame-rate,
fitted in a small TO39 package. Similar to a time-of-flight camera, a thermal camera doesn't rely on environmental
lighting conditions and can easily distinguish background temperature from foreground body temperature, making a
thermal sensor an interesting alternative for this use case.
However, no valid business case can be made for a recognition system with a cheap sensor that requires an expensive processing
platform to host the recognition software. We therefore require an algorithm that fits into low-cost hardware like
microcontrollers.
Using the successful deep learning technology, we propose a light-weight hand-gesture detection framework. 
Besides its limited size and computational complexity, our
proposed model features excellent low-latency properties which are required for this kind of
applications, especially given the low frame-rates that thermal sensors have.
The main contributions of this paper are the following. First, we recorded a new dataset (that has been made
publicly available) with multiple sensors at two viewpoints, including our target sensor MLX90640 and the even cheaper
sensor MLX90641 (with a resolution of 16$\times$12 pixels). Second, a recognition algorithm is proposed that uses a 1D
\textit{Temporal Convolution Network} (TCN) on top of a 2D spatial feature extractor CNN depicted in Figure
\ref{fig:method_overview}. On top of that, we introduce a novelty into the design of our TCN which uses a combination
of \textit{causal} and regular convolutions to drastically improve the low-latency behaviour.
As far as we know, this hasn't been described in literature before. Our algorithm achieves top notch low-latency
detection accuracy when compared to other state-of-the-art models with listed computational complexity and model size.
Finally, we present an empirical study on the influence of the sensor quality and viewpoint on the recognition
performance, where we compare our algorithm on three different thermal sensors.
Our source code is publicly available\footnote{\url{https://gitlab.com/EAVISE/hand-gesture-recognition}}.

Section \ref{sec:related_work} discusses our related work while Section \ref{sec:approach}
proposes our dataset and algorithm. The experiments are detailed in Section \ref{sec:experiments}
and a conclusion follows in Section \ref{sec:conclusion}.

\section{Related work}
\label{sec:related_work}
\noindent {\bf Action recognition:}
Inspired by the big success of deep learning based image analysis like object classification and detection
\cite{girshick2014rich, krizhevsky2012imagenet}, many ground braking video analysis network
designs have been proposed in fields like action/activity recognition
\cite{carreira2017quo, donahue2015long, simonyan2014two, tran2015learning}.
Initial works proposed video recognition algorithms with 2D Convolutions as their core components.
Karpathy et al. \cite{karpathy2014large} compares early, late and slow temporal fusion ConvNet models that
process the frames in short video clips.
Simonian et. al. \cite{simonyan2014two} decomposes an input video into spatial and temporal components. Their
two-stream network design processes the spatial (full frame) and temporal (dense optical flow) components
with individual 2D CNNs. Score fusion merges the output features of both CNNs to get the final result.
Other network designs \cite{carreira2017quo, feichtenhofer2016convolutional, wang2016temporal} exploit the
two-stream concept to further improve upon the state-of-the-art.
A number of works tackle video analysis with sequence modeling
\cite{donahue2015long, sharma2015action, yue2015beyond}. They propose a 2D CNN to capture the spatial context
of each frame into an embedding. These embeddings are sent to an LSTM \cite{hochreiter1997long} which handles
global temporal modeling for activity recognition.
For sequence modeling in general, the Transformer network \cite{vaswani2017attention} and the 1D
Temporal Convolution Network or TCN \cite{bai2018empirical} both report improved performance over recurrent
neural networks like the LSTM. Kozlov et al. \cite{Kozlov2019a} and Farha et al. \cite{farha2019ms}
successfully apply the Transformer and 1D temporal convolution concepts respectively as an alternative for
the LSTM in action recognition. Both report superior performance.
Since a video clip can be seen as a 3D spatio-temporal feature map, many state-of-the-art works have exploited
processing video clips with 3D Convolutions \cite{hara2018can, tran2015learning, tran2017convnet}. The big
advantage of 3D ConvNets is that they can model complex spatio-temporal information in all layers of the
network, which often results in superior performance.
A comprehensive study \cite{carreira2017quo} shows that combining the two-stream approach with 3D networks
leads to an even better result. The downside of 3D CNNs however is that they have a large parameter count
and high computational complexity.

\noindent {\bf Hand gesture recognition:}
Since hand gesture recognition is a form of action recognition, many gesture recognition works use similar
network architectures. Molchanov et al. \cite{Molchanov2016} use a recurrent neural network on top of the C3D
\cite{tran2015learning} feature extractor and test this on color and depth modalities.
They train with CTC loss \cite{graves2006connectionist} which enables the network to learn the start and the
end of a gesture using only a sequence of labels as ground truth without any timing information. On top of that,
their network is able to distinguishing multiple densely concatenated gestures better. We will also use CTC loss
in this work for the same reason. Kopuklu et al. \cite{Kopuklu2019} use a 3D ConvNet gesture detector and a 3D
ConvNet classifier that only activates if the detector has detected a gesture. Besides their good results,
they require a lot of processing power to meet their real-time requirements.
Tateno et al. \cite{tateno2019hand} detect hand gestures on the same thermal sensor as ours.
They use a simple 2D CNN image classifier on top of a background subtracted image. Their background subtraction
however assumes that the hand is always warmer than the clothed body and the background. In practice,
people can have cold hands which makes this method prone to errors. Their dataset is also kept private.

Given the computational complexity and the large amount of parameters needed for 3D CNNs \cite{Kozlov2019a},
we base our lightweight network design on a 2D CNN with a TCN on top as given in Figure
\ref{fig:method_overview}. This design is much more suitable for low-cost hardware and allows
low-latency detection for sensors with low frame-rates like thermal cameras.
We avoid using a two-stream design since this requires duplicating the network resulting in larger models.

\section{Approach}
\label{sec:approach}
After a description of our dataset (Section \ref{sec:dataset}), we discuss the proposed network architecture and how
we train it.

\subsection{Dataset}
\label{sec:dataset}
\begin{figure}
    \centering
    \includegraphics[width=\columnwidth]{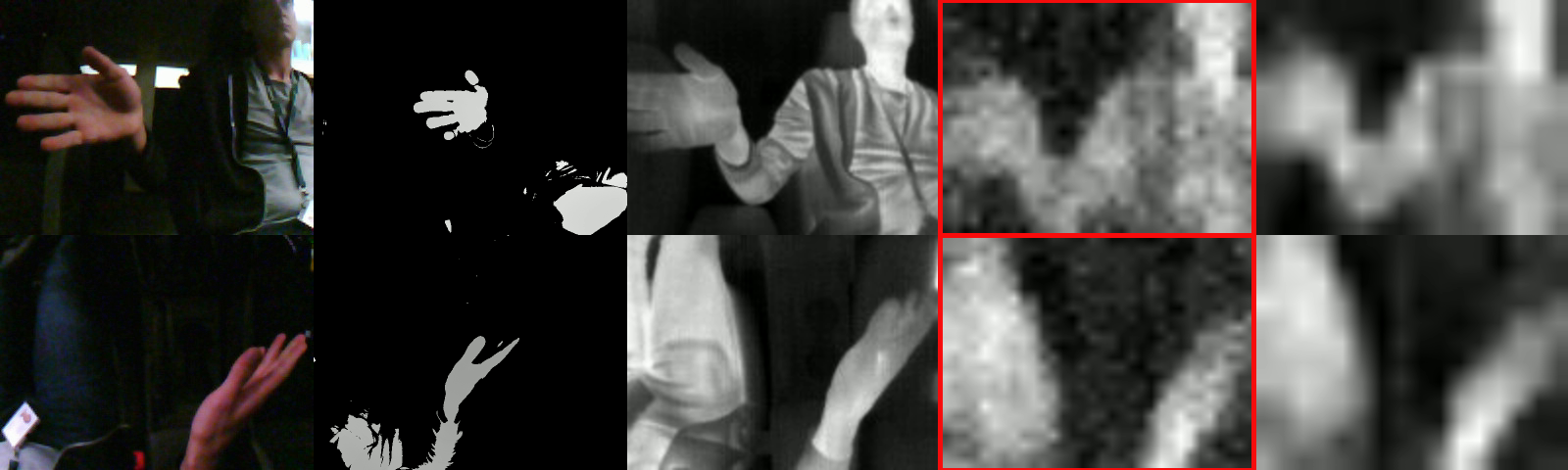}
    \caption{Sample frames from the front sensor cluster (top row) and ceiling sensor cluster (bottom row).
             From left to right: OpenMV color cam, MLX75027 time-of-flight, FLIR lepton, our target sensor
             MLX90640 and MLX90641.}
    \vspace{-1em}
    \label{fig:different_sensors}
\end{figure}

\begin{figure}
    \centering
    \includegraphics[width=\columnwidth]{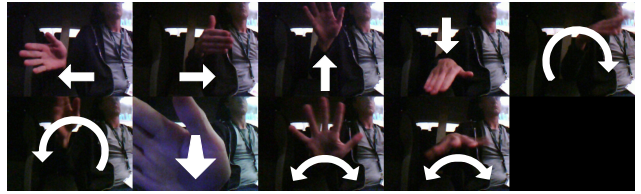}
    \caption{The 9 different hand gestures utilized in our dataset}
    \label{fig:different_gestures}
\end{figure}

Since there are no publicly available hand-gesture recognition datasets with the MLX90640 or MLX90641 sensor, we
recorded a new dataset containing over 1300 hand gesture videos in a car. In order to investigate the influence of the
sensor type and viewpoint, we use two sensor clusters with 5 different cameras each listed in Table \ref{tab:cameras}.
One sensor cluster is mounted on the center of the dashboard in front of the driver, the other is mounted on the
ceiling pointing straight down. The MLX90640 and MLX90641 sensors have a configurable frame-rate between 0.5 and
64FPS. We find 16FPS to be a good trade-off between a decent temporal resolution and spatial noise, which is worse
for higher frame-rates.
Figure \ref{fig:different_sensors} gives some sample frames from the different sensors.
Video labeling is done automatically in a similar fashion as the NvGesture dataset \cite{Molchanov2016};
a digital interface asks the user to execute a predefined hand gesture while it starts recording for 3
seconds. Our dataset contains 9 different dynamic hand gestures from 24 different subjects. The different
gesture classes are shown in Figure \ref{fig:different_gestures} and are especially selected to work with low
and ultra low-resolution sensors. We also recorded a number of \textit{non-gesture} videos which contain actions
like steering, switching gears, operating the radio and operating the wipers. While most gesture videos only
contain a single hand gesture, a number of gesture videos contain two consecutive gestures of the same class.
Our dataset is made publicly available\footnote{\url{https://iiw.kuleuven.be/onderzoek/eavise/mlgesture/home}}.
Due to privacy constraints, only the thermal sensor modalities have been released.

\begin{table}
    \centering
    \adjustbox{width=\columnwidth}{
    \begin{tabular}{l|c|c|c}
    \hline
    type            & modality  & resolution        & FPS (Hz)  \\
    \hline
    OpenMV Cam H7   & Color     & 320$\times$240    & 12        \\
    MLX75027 (ToF)  & Depth     & 640$\times$480    & 16        \\
    FLIR lepton     & Thermal   & 160$\times$120    & 8         \\
    MLX90640        & Thermal   & 32$\times$24      & 16        \\
    MLX90641        & Thermal   & 16$\times$12      & 16        \\
    \hline
    \end{tabular}}
    \caption{Cameras in each sensor cluster}
    \vspace{-1em}
    \label{tab:cameras}
\end{table}

\subsection{Network architecture}
\label{sec:network_arch}
\begin{figure}
    \centering
    \includegraphics[width=.6\columnwidth]{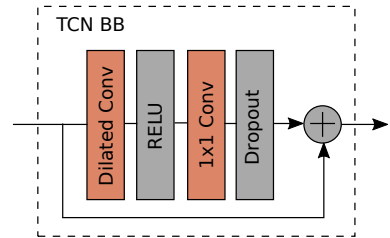}
    \caption{TCN basic block}
    \vspace{-1em}
    \label{fig:tcn_bb}
\end{figure}

\begin{figure*}
    \centering
    \includegraphics[width=\textwidth]{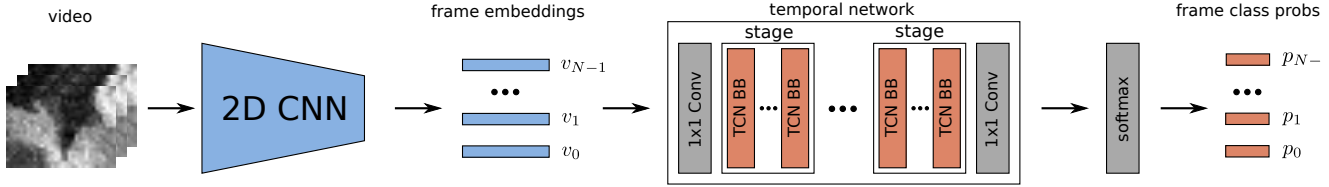}
    \caption{Proposed generic network architecture}
    \label{fig:network_arch}
\end{figure*}

Our proposed network architecture, illustrated in Figure \ref{fig:network_arch}, uses a 2D CNN to create an
embedding vector $v_n\in\mathbb{R}^C$ for each video frame and models the temporal domain using a 1D
\textit{Temporal Convolution Network} (TCN) based on the work of Farha et. al. \cite{farha2019ms}.
We initially use ResNet18 \cite{he2016deep} as our spatial encoder CNN.

Many works \cite{oord2016wavenet, bai2018empirical, farha2019ms, wan2019multivariate}
advertise the superior properties of TCNs compared to recurrent networks such as better accuracy, smaller
model sizes and high modularity. We use the same TCN basic block as proposed by Farha et. al.
\cite{farha2019ms}, depicted in Figure \ref{fig:tcn_bb}. From now on, we will refer to this basic block as
BB. The first layer of the BB is a 1D dilated convolution with kernel size $k=3$ where the
dilation factor is doubled for each subsequent BB, i.e. 1, 2, 4, 8,... This allows the
receptive field to grow exponentially.
A stack of subsequent BBs is called a \textit{stage}. To create deeper networks, multiple \textit{stages} can be
stacked on top of each other where the dilation factor is reset to 1 at the start of every new stage.
As in \cite{farha2019ms}, no down-sampling is used, so the temporal output resolution is identical to the temporal
input resolution.

\begin{figure}
    \begin{subfigure}[b]{\columnwidth}
        \includegraphics[width=\columnwidth]{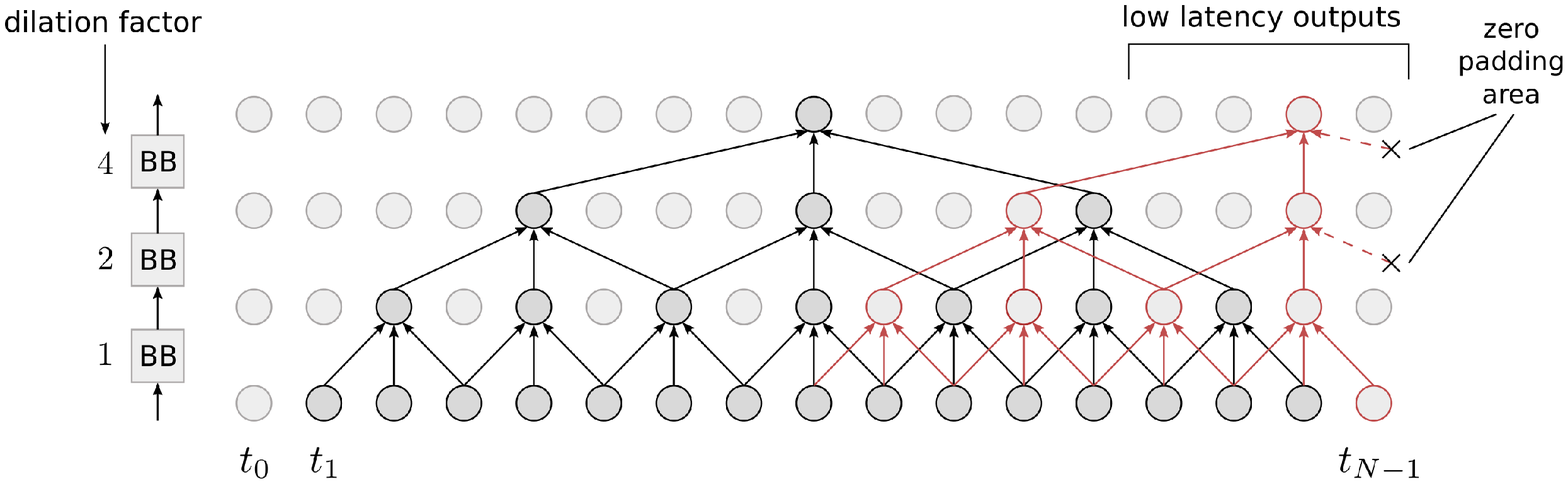}
        \caption{Non-causal convolutions}
        \label{fig:non_causal}
    \end{subfigure}
    \break
    \begin{subfigure}[b]{\columnwidth}
        \includegraphics[width=\columnwidth]{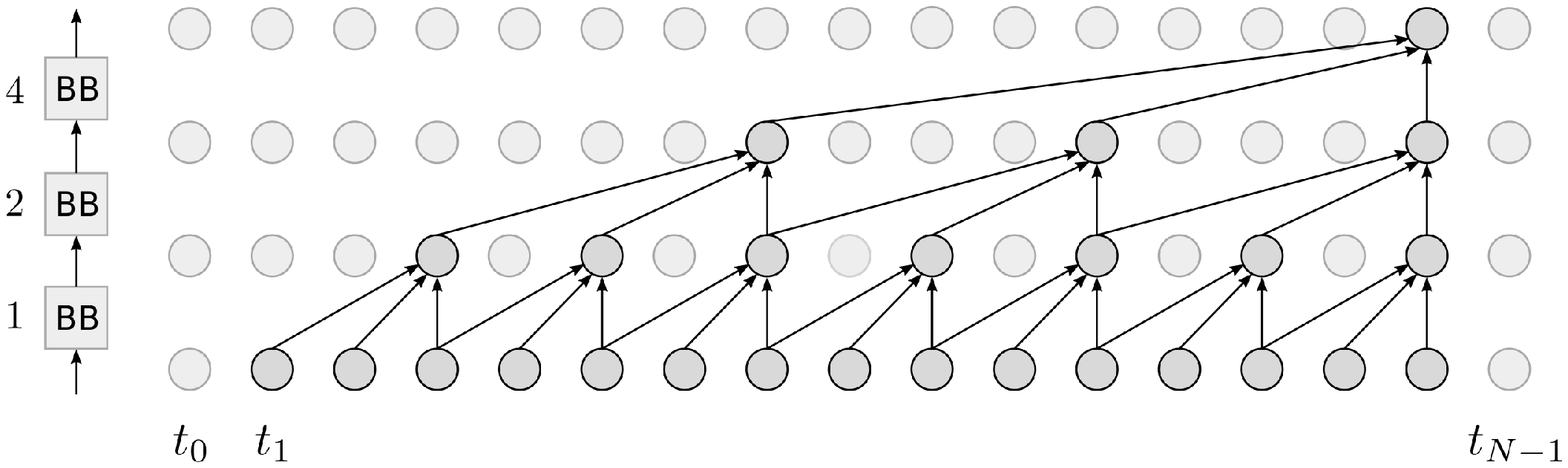}
        \caption{Causal convolutions}
        \label{fig:causal}
    \end{subfigure}
    \break
    \begin{subfigure}[b]{\columnwidth}
        \includegraphics[width=\columnwidth]{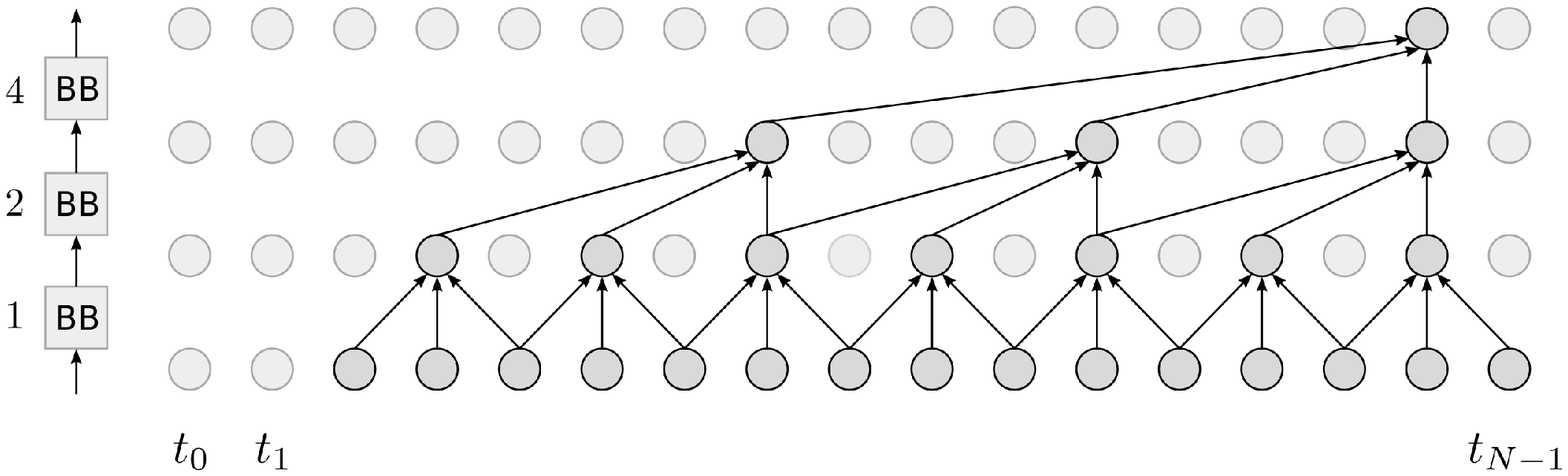}
        \caption{Mixed causal convolutions}
        \label{fig:mixed_causal}
    \end{subfigure}
    \caption{Comparison of non-causal, causal and mixed-causal TCNs}
\end{figure}

Figure \ref{fig:non_causal} gives an illustration of the receptive field (black arrows) of an output activation
from a single stage with three stacked BBs. This example uses regular convolutions as proposed by Faraha et. al. 
\cite{farha2019ms}. Regular convolutions have a receptive field that expands equally wide to the right as it does
to the left. For a temporal convolution, this means that it looks as far into the \textit{future} (right) as it 
looks into the \textit{past} (left).
However, output activations close to the right edge of the network's temporal window,
e.g. the one depicted in red in Figure \ref{fig:non_causal}, are mostly produced by left and down looking kernel 
elements since most right looking kernel elements receive data from the zero padding area. This results in poor 
recognition performance at the edges compared to activation outputs at the center, especially since the receptive
field is significantly large due to the exponentially increasing dilations. We also prove this in the experiments 
section. This is bad for low latency applications, since they cannot afford looking (far) into the future and thus 
require predictions close to the right edge of the temporal window. To boost the performance for predictions close to 
the right edge, one solution is to use \textit{causal} convolutions \cite{bai2018empirical, oord2016wavenet}, which
only look at present and past activations (Figure \ref{fig:causal}).
We however empirically discovered that a TCN with \textit{causal} convolutions performs below par compared to a TCN
with regular convolutions, when considering central located activation outputs.
Also, since low-latency doesn't necessarily mean zero latency, a few future samples
can still be used. We therefore propose using a mixture of BBs with a regular (a.k.a.
\textit{non-causal}) dilated convolution and BBs with a \textit{causal} dilated convolution to become an asymmetric 
receptive field that only looks a few frames ahead and a multitude of frames back. We coin this configuration 
\textit{mixed causal}.
Figure \ref{fig:mixed_causal} illustrates a \textit{mixed causal} configuration where the first BB is \textit{non-causal}
and allowed to look ahead while the other BBs are \textit{causal}. We compare different \textit{mixed causal} 
configurations in the experiment section and empirically prove their superior performance for low-latency applications.

The input of the stacked TCN stages is constructed from a sequence of embedding vectors $\{v_0, v_1,...,v_N\}$
coming from the 2D CNN (Figure \ref{fig:network_arch}). The sequence is concatenated into a 2D feature map 
$\pmb{v}\in\mathbb{R}^{C\times N}$ with $C$ the number of feature channels and $N$ the length of temporal sequence.
A $1\times1$ convolution compresses the number of feature channels $C$ to $C'$, the number of feature channels used
inside the BBs. The output feature map $\pmb{y}\in\mathbb{R}^{C'\times N}$ of the last BB is 
finally converted to frame-wise class probabilities $\{p_0, p_1, ..., p_N\}$ with $p_n\in\mathbb{R}^P$ where $P$ is the 
number of supported classes including the \textit{non-gesture} class, through another 1$\times$1 convolution and a 
softmax layer.

\subsection{Loss functions}

Our dataset includes per-video labels rather than time segmented annotations, since this way, annotation can be done
automatically, as explained in Section \ref{sec:dataset}. With video labels, we can
easily use cross-entropy loss (CE loss) for offline video classification purposes. This can be done by averaging
all logits of each temporal output step before sending the result through the softmax layer and thus becoming a
single classifier output.
However, since we require low-latency predictions that can distinguish rapid consecutive hand gestures,
training our network on a temporal segmentation task would be more desirable. Temporal segmentation however
requires frame-wise annotations. To avoid manual annotation work for marking the nucleus of each gesture in
the training set, we can solve this problem by using CTC loss, short for
\textit{Connectionist Temporal Classification} loss \cite{graves2006connectionist}.
CTC loss can induce frame-wise gradients, given only a target sequence without timing information. This way,
our proposed network can be trained for segmentation-like tasks without
frame-wise annotations.
We can easily deduce target sequences from video class labels using the following scheme: A video with the
\textit{non-gesture} class label gets an empty target sequence $\{\}$, a video with a hand gesture label $l$ gets a
target sequence $\{l\}$ and a video with two consecutive hand gestures of the same class gets target sequence
$\{l, l\}$. Note that we don't need to reproduce this timeless target sequence during inference, which allows
us to avoid using any form of sequence decoding like best-path decoding or beam-search decoding in our final
application. Instead, a simple threshold on the temporal segmentation-like output probabilities of one of the
network's output time steps is sufficient. Figure \ref{fig:wave_forms} gives an example of output
probabilities produced by a network trained on a classification task using cross-entropy loss and a network
trained with CTC loss.
Molchaniv et. al \cite{Molchanov2016} also use this idea for their hand gesture recognition network.
We elaborate further on this in the experiments section.

\subsection{Training details}
We use a 2D CNN with a 32$\times$32 input resolution and initialize it with pre-trained weights from the CIFAR-10
dataset \cite{krizhevsky2009learning}. The pre-trained weights of the first convolution layer that process RGB
channel images are averaged to support single channel thermal images. The temporal network gets initialized with random
parameters. We then train the model with CE loss and fine-tune it later with CTC loss on the best performing CE-trained
weights.

Before sending the calibrated temperatures from the thermal sensors to the network, we first normalize them
to ensure that the distribution of the input data matches the expected
distribution of the 2D CNN's pre-trained CIFAR-10 weights ($\mu = 0, \sigma = 1$).
We use data augmentation in addition to avoid over-fitting. Next to random corner cropping, we use random
contrast and brightness augmentation for all thermal sensors to make the models robust against changes in
hand, body and background temperatures. For the thermal MLX sensors, we also add additional Gaussian
distributed noise with random spread levels. This discourages the model to try to fit on the noise embedded in
the images from the dataset. We empirically found that this gives better results. We also apply temporal
shifting and scaling of 25\% and 20\% respectively.

We use a randomly selected train and test dataset split of 70\% and 30\% respectively.
Training is done for 150 epochs with a batch size of 8 and the Adam optimiser is used with a learning rate
of 1e-4. The learning rate is lowered by a factor of $\times$10 when the test accuracy reaches a
plateau of 20 epochs. Both CE and CTC loss training schedules use the same settings.

\section{Experiments}
\label{sec:experiments}
We first evaluate our proposed network architecture multiple times as a video
classifier, in search for the best hyper parameters of the temporal network.
We then compare our architecture to other possible classification designs.
In these experiments a single prediction per video is requested from a model. To evaluate the
low-latency detection performance, we run our model and its competitors in a sliding-window based fashion
over a long video, evaluating the generated per-frame predictions. These experiments are performed on the MLX90640
sensor data coming from the sensor cluster on the dashboard. In Section \ref{sec:other_sensor_results},
a performance comparison is presented among networks trained on the other thermal sensor types and sensor
viewpoints available in our dataset.

\subsection{Evaluation metrics}

To measure the video classification accuracies, we calculate the top-1 classification scores by running
the network on each video of the test set.
We report the performance for each model when trained with CE loss and CTC loss.
For networks trained with CTC loss, we use the following scheme to produce a video classification label:
First, best-path decoding \cite{graves2006connectionist} is used to get the predicted sequence of class
labels. We then use the label within the decoded sequence that has the highest probability score as our
final class label. This works since all videos in our dataset have either none, one or two gestures of the
same class label.
All classification experiments are run three times with different manual random seeds. We
report the average results.

To measure the detection performance in time, we first stitch 50\% of all test set videos together in one
long duration test video. We then manually annotate the hand gesture nuclei in the resulting
test video and run the network in a sliding-window based fashion over all frames in the video. A single
temporal output step of the network produces the prediction output as depicted in Figure
\ref{fig:method_overview}.
In Section \ref{sec:low_latency_experiments}, we will empirically determine which output is most suited
with the help of the following detection metric.
We calculate a \textit{Precision-Recall} (PR) curve for each class label, except for the
\textit{non-gesture} label, and average all curves to a multi-class PR-curve. The
\textit{mean Average Precision} or mAP, which is defined as the area under the final PR-curve, is used as our
detection metric.
If the network fires once during an annotated gesture nucleus, it is considered a True
Positive. A second detection during the nucleus or detections outside the nucleus are considered False
Positives. Missed gestures are considered False Negatives. We prefer the PR-curve over the ROC curve
\cite{Molchanov2016} since defining the correct amount of True Negatives, which is required to calculate
the ROC curve, is too difficult given the spiky nature of the CTC predictions.

\subsection{Hyperparameters proposed TCN}
\label{sec:hyper_params}

In these experiments, we vary the structure of our proposed TCN network to search for an optimal performing solution
using the MLX90640 low-resolution thermal data, taken from the sensor cluster located on the dashboard. First, a grid
search is done with only \textit{non-causal} convolutions where we train networks with 4, 5 and 6 BBs
for each 1, 2 and 4 stages. This experiment is done with $C'=64$ feature channels in each BB and repeated with
$C'=128$ feature channels afterwards. We find that 4 stages with 5 BBs works best on both 64-channel and
128-channel experiments. These results are consistent when training with CE and CTC loss.

\subsection{Comparison with other methods}

\begin{table*}
\centering
\adjustbox{width=.9\textwidth}{
\begin{tabular}{lcccccccc}
\toprule
              Model & CE Acc &  CE Gap & CTC Acc & CTC Gap & FLOPS TN & Params TN &    FLOPS &  Params \\
\midrule
           baseline &  79.7\% &  -16.7\% &   42.4\% &   -6.1\% &       - &        - &  556.27M  &   11.17M \\
3D CNN \cite{hara2018can} &   95.4\% &   -2.6\% &        - &    - &    - &        - &    9.75G  &  33.15M \\
                GRU &  94.0\% &   -3.2\% &   89.0\% &   -5.3\% &   76.04M &     1.58M &  632.06M &  12.75M \\
             BI-GRU &  94.7\% &   -2.9\% &   93.9\% &   -3.3\% &  152.08M &     3.16M &  708.10M &  14.33M \\
               LSTM &  94.7\% &   -1.2\% &   85.3\% &   -5.5\% &  101.30M &     2.11M &  657.33M &  13.27M \\
            BI-LSTM &  94.8\% &   -2.6\% &   93.4\% &   -1.6\% &  202.61M &     4.21M &  758.63M &  15.38M \\
           GRU attn &  94.5\% &   -2.1\% &   85.5\% &   -5.8\% &   76.04M &     1.58M &  632.06M &  12.75M \\
          LSTM attn \cite{sharma2015action} &  93.6\% &   -3.4\% &   91.6\% &   -3.2\% &  101.30M &     2.11M &  657.33M &  13.27M \\
        BI-GRU attn &  94.6\% &   -2.7\% &   92.2\% &   -2.2\% &  152.08M &     3.16M &  708.10M &  14.33M \\
       BI-LSTM attn &  94.3\% &   -1.2\% &   91.4\% &   -5.0\% &  202.61M &     4.21M &  758.63M &  15.38M \\
                VTN \cite{Kozlov2019a} &  91.4\% &   -6.6\% &   76.9\% &  -11.1\% &   58.37M &     1.84M &  614.39M &  13.01M \\
\midrule
           TCN f128 (ours) &  \bf{95.9\%} & -1.2\% &  \bf{95.2\%} &   \-0.4\% &   66.37M &     1.38M &  622.40M &  12.55M \\
            TCN f64 (ours) &  94.6\% &   -1.8\% &   94.1\% &   -0.6\% &   \bf{17.46M} &     \bf{0.36M} &  573.48M &  11.53M \\
    TCN f64 SqueezeNet (ours) &  93.0\% &    \bf{1.3\%} &   91.7\% &    \bf{1.4}\% &   \bf{17.46M} &     \bf{0.36M} &   \bf{34.01M} &   \bf{1.09M} \\
\bottomrule
\end{tabular}}
\caption{Results of competitive video classifier designs. Next to \textit{CE Acc} and \textit{CTC Acc} trained network accuracies,
         \textit{CE Gap} and \textit{CTC Gap} represent the test accuracy minus the training accuracy. \textit{FLOPS TN}
         and \textit{Params TN} represent the number of FLOPS and parameters of the \textit{Temporal Network} only
         (for processing a sequence of 48 time steps) while \textit{FLOPS} and \textit{Params} represent the number of FLOPS
         and parameters of the whole network.}
\vspace{-1em}
\label{tab:classification_results}
\end{table*}

In this section, we analyse the performance of our proposed architecture against other network designs.
Table \ref{tab:classification_results} gives the results of different models trained on input data from the
MLX90640 sensor on the car's dashboard. We use ResNet18 as the default spatial encoder and set the temporal length to
48 frames, which for the MLX90640 approximates to the full length of each test video. We also tried the larger
ResNet34, but found it to perform worse on our dataset.
We present our proposed model with 4 stages and 5 BBs in both 64 feature channels and 128 feature channels
configuration, coined \textit{TCN f64} and \textit{TCN f128} respectively. \textit{TCN f64 SqueezeNet} is our 
\textit{TCN f64} model with a SqueezeNet V1.1 spatial encoder \cite{iandola2016squeezenet} instead of ResNet18, to put 
into contrast the additional reduction in FLOPS and number of parameters that can be achieved by reducing the model 
capacity of the spatial encoder.
All RNNs are single layer and retrained a second time with a spatial attention module \cite{sharma2015action},
resulting in models denoted with an \textit{attn} postfix, in the hope that the spatial context could be captured
better. The \textit{Video-Transformer Network} or VTN \cite{Kozlov2019a} uses the proposed configuration from its
publication. To put the results in perspective, a model trained \textit{without} a temporal network (containing only
a spatial encoder) coined \textit{baseline} and a 3D-inflated ResNet18 coined \textit{3D CNN} \cite{hara2018can}
are also presented.

The \textit{baseline} model has the lowest accuracy for both \textit{CE} and \textit{CTC} trained
models, which clearly indicates the importance of the temporal network component.
Both our \textit{TCN f64} and \textit{TCN f128} models outperform all other sequence modeling designs when trained
with CTC-loss, which for detection is the most important metric. Also the difference between test and train
accuracy, listed by the \textit{CTC Gap} column, is lowest for our proposed models, indicating less chance of
over-fitting compared to the other models.
\textit{TCN f128} even outperforms the \textit{3D CNN} model slightly. Since the 3D CNN only features a single
classifier output in time, it cannot be trained with CTC loss.

In the last four columns, the number of theoretical FLOPS and parameters are presented for the temporal network
only and the whole network respectively.
The number of parameters listed under \textit{Params TN} for our proposed TCNs is less and even far less
for \textit{TCN f64} compared to the recurrent models, which is beneficial for devices with smaller
high-bandwidth memory sizes.
Counting the number of FLOPS is dependent on the mode of operation.
In our case, the network runs in a sliding-window based fashion over a video feed, generating a new prediction
immediately after processing a new frame.
Since already calculated embedding vectors from the previous frames can be stored in memory, the \textit{FLOPS}
column lists the FLOPS for executing the spatial encoder once plus the
FLOPS to process the 48 most recent embeddings with the temporal network. \textit{FLOPS TN} reports the lowest
count for our \textit{TCN f64} model. Since ResNet18 requires the biggest portion of FLOPS and parameters,
modifying the spatial encoder architecture is the first thing to do when further optimization is required.
Our SqueezeNet variant clearly shows that future experiments can drastically reduce the overall network size
if needed.

\subsection{Mixed causal configurations}

\textit{Mixed causal} configurations, as presented in Section \ref{sec:network_arch}, provide a nice way to
boost the detection accuracy of low-latency network outputs, which is discussed in more detail in section
\ref{sec:low_latency_experiments}.
To find good mixture configurations, we redo the grid search from Section \ref{sec:hyper_params} but now
with different \textit{mixed causal} configurations.
To limit the number of possible combinations, we only try networks with 2 and 4
stages and make sure that each stage in the same model has an identical causal configuration.
We vary the number of \textit{non-causal} BBs at the beginning of each stage and train networks with
1, 2 and 3 \textit{non-causal} BBs per stage while all higher BBs in a stage are set to
\textit{causal}.
We find that 4 stages and 4 BBs gives the best result for any \textit{mixed causal} configuration. Further
results for \textit{mixed causal} models are therefore reported with this hyper parameter set.

Table \ref{tab:mixed_causal} lists the CTC-trained network classification results of different
causal mixtures for the \textit{TCN f128} model, together with its already trained \textit{non-causal} variant
and a pure \textit{causal} variant.

\begin{table}
\centering
\adjustbox{width=\columnwidth}{
\begin{tabular}{lccccc}
\toprule
           Model & CTC Acc & CTC Gap &  \#non-causal &  \#causal \\
\midrule
        TCN f128 &   95.2\% &   -0.4\% &            5 &        0 & \\
 TCN causal f128 &   90.5\% &   -2.6\% &            0 &        5 & \\
   TCN mix1 f128 &   92.7\% &   -2.3\% &            1 &        3 & \\
   TCN mix2 f128 &   93.9\% &   -1.9\% &            2 &        2 & \\
   TCN mix3 f128 &   93.8\% &   -0.1\% &            3 &        1 & \\
\bottomrule
\end{tabular}}
\caption{Non-causal, causal and mixed causal configuration accuracies. Columns \textit{\#non-causal} and
         \textit{\#causal} list the number of \textit{non-causal} and \textit{causal} BBs per stage
         respectively.}
\vspace{-1em}
\label{tab:mixed_causal}
\end{table}

The classification results of the mixed models in Table \ref{tab:mixed_causal} are lower compared to their
\textit{non-causal} variant. This makes sense since a \textit{non-causal} network looks further into the
future than a mixed model if it is allowed to, which is the case since this classification test provides
all video frames at once before a decision needs to be made.
However, when no or only a few future frames are available when a prediction needs to be made,
say in a low-latency recognition task, the recognition performance of a \textit{non-causal} TCN will drop
drastically. This is where our mixture models prevail. Section \ref{sec:low_latency_experiments} explains
this behaviour for low-latency detection.

\subsection{Low-latency detection performance}
\label{sec:low_latency_experiments}
\begin{figure*}
    \centering
    \includegraphics[width=.72\textwidth]{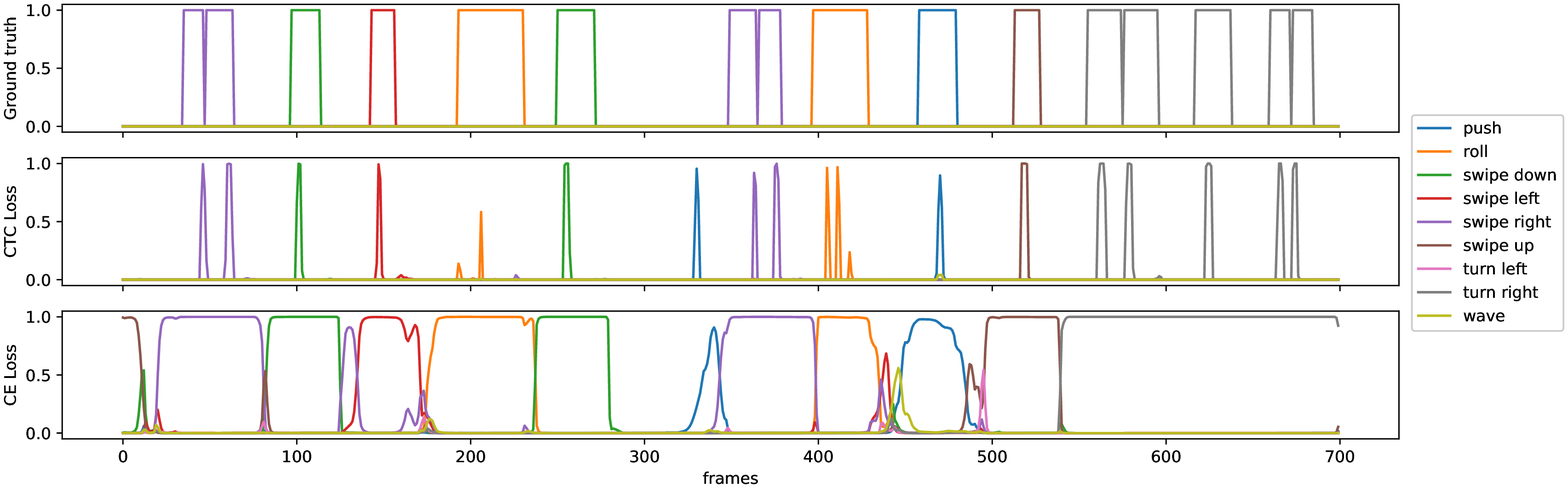}
    \vspace{-1em}
    \caption{Network output probabilities from a CTC and CE trained network versus the ground truth.}
    \label{fig:wave_forms}
    \vspace{-1em}
\end{figure*}

In this section we analyse the detection performance on the MLX90640 thermal data using the mAP metric.
Figure \ref{fig:wave_forms} gives an example of the annotated ground truth together with the output
predictions of a CTC and CE trained network. The spiky predictions of a CTC-trained network
can clearly distinguish rapid consecutive gestures while the CE trained network tends to merge
predictions of the same class together. All detection results discussed in this section are therefore created
with CTC-trained networks.

\begin{figure*}
    \centering
    \begin{subfigure}[b]{.3\textwidth}
        \includegraphics[width=\columnwidth]{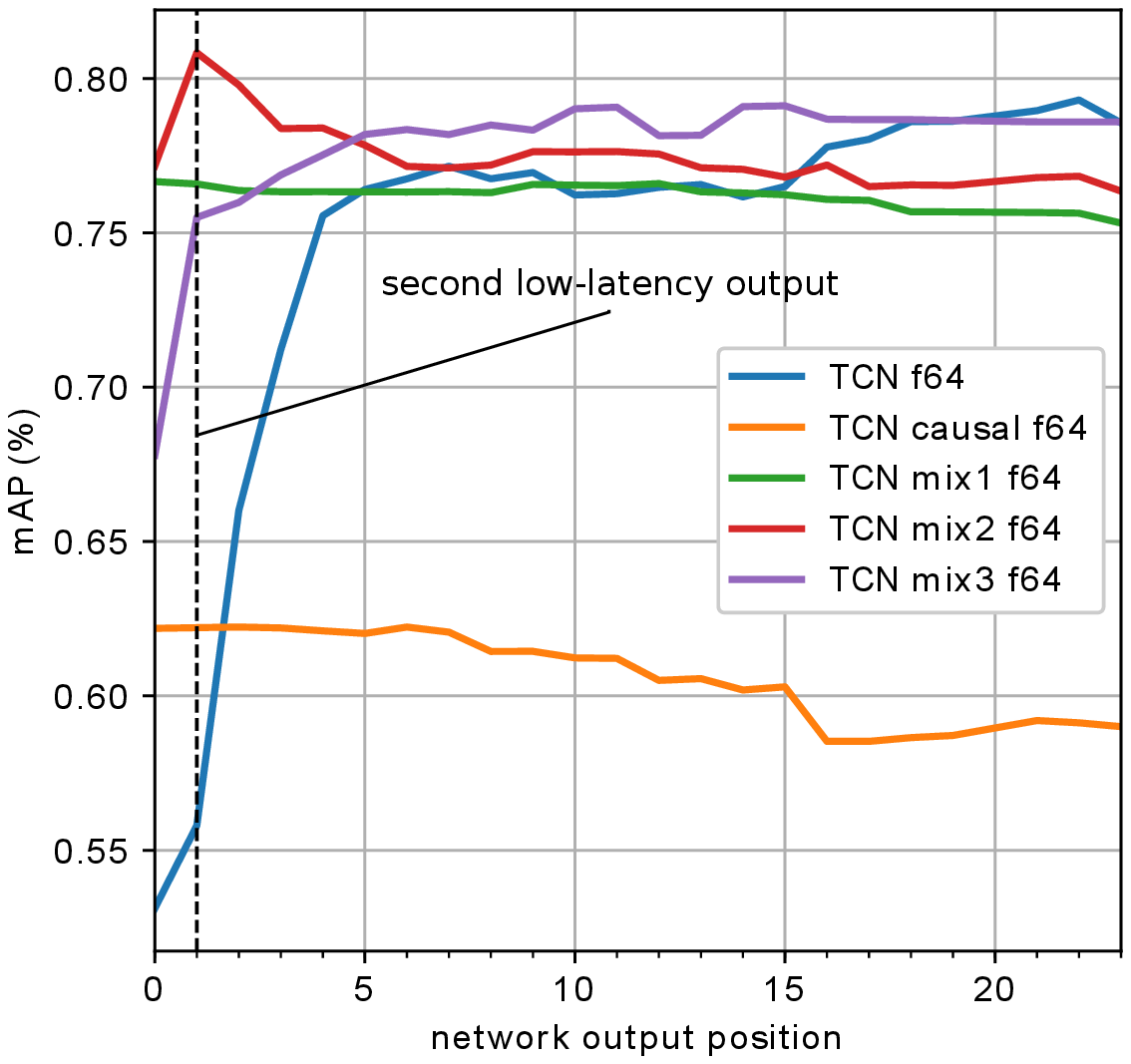}
        \caption{\textit{Mixed causal} 64 feature channels}
        \label{fig:tcn_mixed_f64}
    \end{subfigure}
    \begin{subfigure}[b]{.3\textwidth}
        \includegraphics[width=\columnwidth]{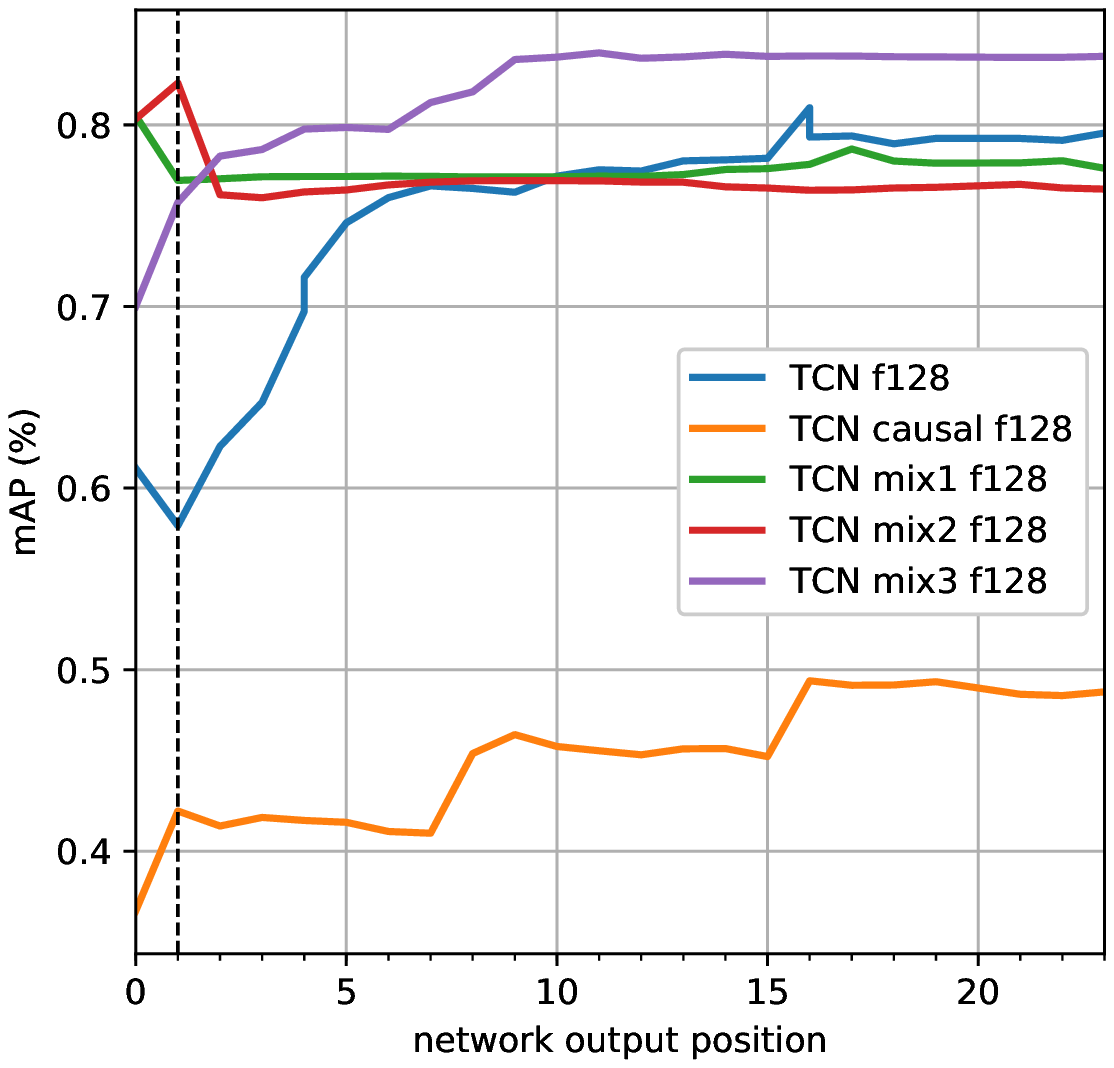}
        \caption{\textit{Mixed causal} 128 feature channels}
        \label{fig:tcn_mixed_f128}
    \end{subfigure}
    \begin{subfigure}[b]{.3\textwidth}
        \includegraphics[width=\columnwidth]{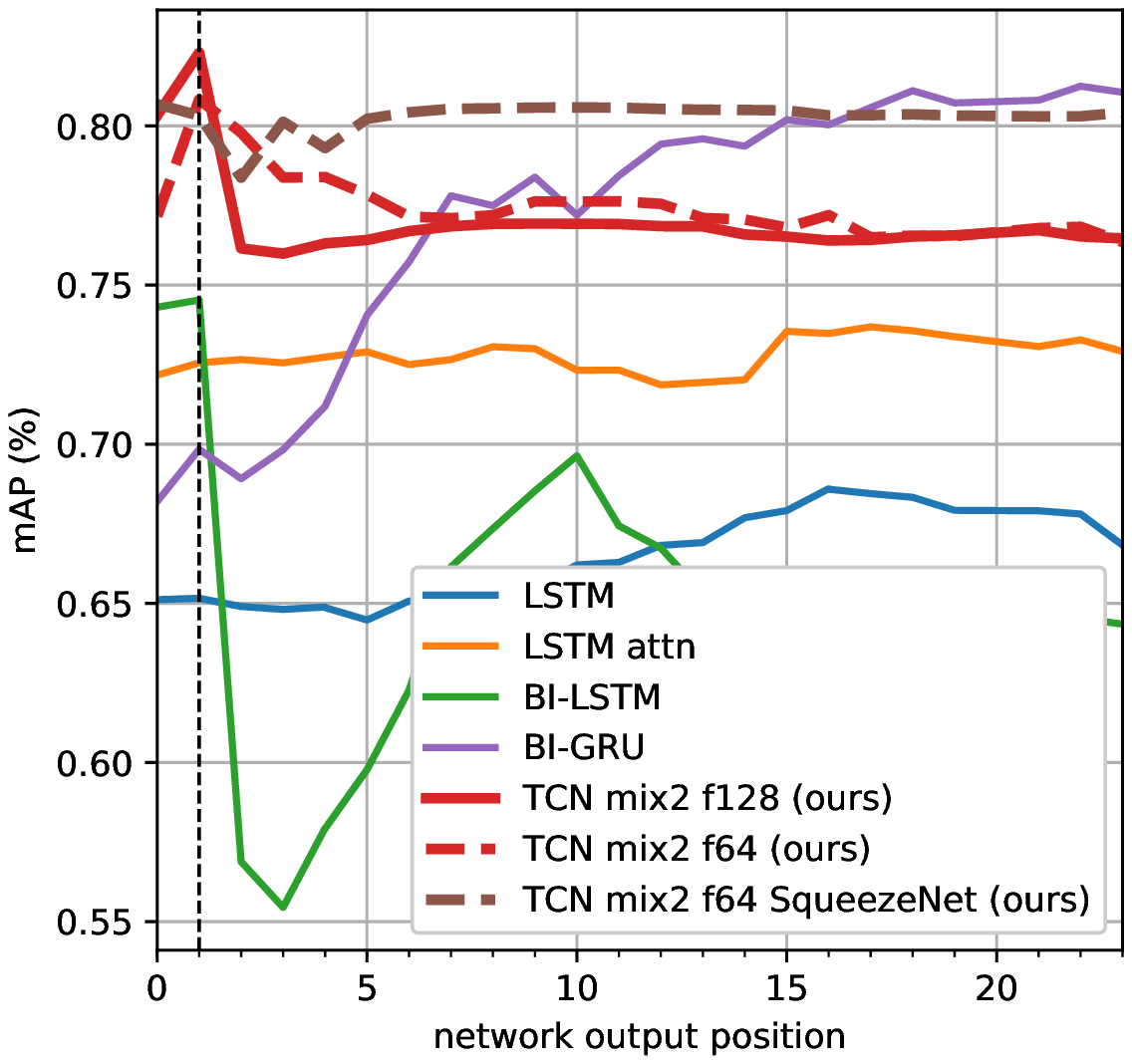}
        \caption{Comparison to other architectures}
        \label{fig:tcn_others}
    \end{subfigure}
    \caption{Low-latency detection performance in \textit{mean Average-Precision} versus network output position
             graphs.}
    \label{fig:detection_results}
    \vspace{-1em}
\end{figure*}

Given the low-latency requirement, the graphs in Figure \ref{fig:detection_results} present the mAP results
versus the network output position, where 0 and 23 on the x-axis represent the index of most right and center
network outputs respectively. In general, high accuracies at the left side of the graph are important for
low-latency applications.

Figure \ref{fig:tcn_mixed_f64} and \ref{fig:tcn_mixed_f128} both report poor performance for the regular
\textit{non-causal} models on the low-latency output region of the graph compared to the more central located
outputs, clearly indicating the issue discussed in Section \ref{sec:network_arch}. The vanilla \textit{causal}
models do not suffer from this effect, but since they don't learn from any near future inputs, their performance
is below par. The right mixture of low-level \textit{non-causal} and high level \textit{causal} BBs however solves 
this problem with outstanding low-latency detection performance. The best network output 
positions of \textit{mix2} and \textit{mix3} even outperform the best network output positions of the 
\textit{non-causal} models, making our mixture strategy the best choice. \textit{mix2}
provides us with a nice peak on the second low-latency output in both Figure \ref{fig:tcn_mixed_f64} and
\ref{fig:tcn_mixed_f128}. Using this output, we can build a processing pipeline with a latency of only a single
frame. With 16 FPS, this boils down to a latency of $1/FPS=1/16=62.5ms$.

Figure \ref{fig:tcn_others} compares the \textit{mix2} models with the detection performance of the best
competitive models from Table \ref{tab:classification_results}. Here, the low-latency detection performance of our
mixture designs crushes that of the best competition. Note that even with the \textit{SqueezeNet} backbone, our TCN
performs outstandingly well. A detection demo video of model \textit{TCN mix2 f128} can be 
found online\footnote{\url{https://www.youtube.com/watch?v=DAjKO0HWW78}}.

\subsection{Other sensors and viewpoints}
\label{sec:other_sensor_results}

To evaluate the effect of sensor quality and viewpoint, we present an empirical study on the classification
performance of our proposed \textit{TCN f128} model, trained on the three thermal sensor types for two
different viewpoints. To avoid optimizing the 2D CNN for even lower resolutions, we upsample the MLX90641
data to 32$\times$24 pixels. We also downsample the FLIR Lepton data to 32$\times$24 pixels to study the
effect of image quality rather than resolution, since the FLIR Lepton provides much cleaner images compared
to the MLX sensors, even on the same resolution.
Table \ref{tab:results_modalities} lists the CE and CTC results. From these results we can conclude that the
\textit{front} models perform slightly better in general compared to the \textit{top} models. Also the image
quality matters, because the FLIR lepton results outperform the MLX90640 results. But maybe the most
interesting remark is that although only half the resolution, the performance of the MLX90641 trained models
rather closely follow the performance of the MLX90640 trained models. This opens opportunities for an even
more affordable recognition system since the MLX90641 is cheaper and only requires half of the pixels to be
processed, which can reduce the computational cost of the spatial encoder even further.

\begin{table}
\vspace{0.5em}
\centering
\adjustbox{width=\columnwidth}{
\begin{tabular}{lcccc}
\toprule
             Model & CE Acc & CE Gap & CTC Acc & CTC Gap \\
\midrule
    MLX90641 front &  97.0\% &  -0.8\% &   92.9\% &   -2.3\% \\
    MLX90640 front &  95.9\% &  -1.2\% &   95.2\% &   -0.4\% \\
 FLIR Lepton front &  98.0\% &  -1.0\% &   96.6\% &   -1.6\% \\
\midrule
      MLX90641 top &  94.0\% &  -3.6\% &   92.9\% &   -2.3\% \\
      MLX90640 top &  94.2\% &  -3.3\% &   94.9\% &   -2.7\% \\
   FLIR Lepton top &  95.9\% &  -3.1\% &   95.2\% &   -2.0\% \\
\bottomrule
\end{tabular}}
\caption{Classification results of different thermal sensors and their viewpoint.
         \textit{Front} means dashboard location while \textit{top} means ceiling location.}
\label{tab:results_modalities}
\vspace{-1em}
\end{table}

\section{Conclusion}
\label{sec:conclusion}
In this work, we investigated the possibility to design a low-cost variant of an
in-car hand-gesture enabled human-machine interface for multiple purposes.
For this, we used a cheap low-resolution thermal camera with only 32$\times$24 
pixels. We proposed a light-weight recognition framework that can be
easily optimized for low-cost hardware and recorded a new dataset of over 1300 
gesture videos to train our models. Our dataset features multiple sensors, two 
viewpoints and has been made publicly available.
Our proposed algorithm uses a temporal convolution
network on top of a 2D CNN, which features excellent low-latency properties thanks 
to our newly proposed technique of mixing \textit{causal} and \textit{non-causal}
convolution layers. We achieved up to 95.9\% classification accuracy and 83\% mAP 
detection accuracy with a latency of only one frame. In a final study we compared
the recognition performance of models trained on three different thermal sensors
types and two viewpoints. These last experiments concluded that sensor quality
influences the prediction result and that a sensor of 16$\times$12 pixels is an
excellent choice for reducing the system's cost even further. 
In future work, possibilities for model compression and quantization to fit
this network into a low-cost microcontroller will be investigated.

\section{Acknowledgements}

This work is supported by VLAIO and Melexis via the Start to Deep Learn TETRA project.

{\small
\bibliographystyle{ieee_fullname}
\bibliography{manuscript}
}

\end{document}